\documentclass[lettersize,journal]{IEEEtran}
\IEEEoverridecommandlockouts                  
\usepackage{graphics} 
\usepackage{graphicx}
\usepackage{balance}
\usepackage{booktabs}
\usepackage{dblfloatfix}
\usepackage{placeins}
\renewcommand{\dbltopfraction}{0.92}
\renewcommand{\dblfloatpagefraction}{0.75}
\renewcommand{\textfraction}{0.06}
\usepackage{amsmath,amsfonts}
\usepackage{amssymb}
\usepackage{algorithmic}
\usepackage{algorithm}
\usepackage{threeparttable}
\usepackage[font=small,labelfont=bf,tableposition=top,hypcap=false]{caption}
\usepackage{array}
\usepackage[caption=false,font=normalsize,labelfont=sf,textfont=sf]{subfig}
\usepackage{textcomp}
\usepackage{url}
\usepackage{verbatim}
\usepackage{capt-of}
\usepackage[compact]{titlesec}

\usepackage{cite}
\usepackage{CJKutf8}
\usepackage{subfig}
\usepackage{float}
\usepackage{amsmath}
\makeatletter
 \let\NAT@parse\undefined
 \makeatother
\usepackage[pdftex,
        colorlinks=true,
        bookmarks=true,
        citecolor=red,
        linkcolor=blue,
        pagebackref=true,
        pdfstartview=Fit,
        breaklinks=true
]{hyperref}
\usepackage{cleveref}
\begin{document}
\newcommand{\blue}[1]{{\color{blue}#1}}
\newcommand{\red}[1]{{\color{red}{\sout{#1}}}}
\newcommand{\etal}{\textit{et al.~}}
\def\etc{\textit{etc.~}}  
\def\eg{\textit{e.g.},~}
\def\ie{\textit{i.e.},~}

\title{\LARGE\bf
Phrase-Level Robotic Guqin Performance: Bimanual Motion Planning and Audio-Tactile Interaction Monitoring}
\author{Zhen Wang$^{1,\dagger}$, Zhiheng Chen$^{1,\dagger}$, Tianyuan Bao$^{1,\dagger}$ and Tianwei Zhang$^{1,3,*}$ 
\thanks{$^{1}$The Shenzhen Institute of Artificial Intelligence and Robotics for Society, Shenzhen, China}
\thanks{{$^2$}The Chinese University of Hong Kong - Shenzhen, Shenzhen, China}
\thanks{$^{\dagger}$These authors contributed equally to this work.}%
\thanks{* Corresponding Author: \tt\small {zhangtianwei@cuhk.edu.cn}}
}
\maketitle
\vspace{-10pt}
\begin{CJK}{UTF8}{gbsn}
\begin{abstract}
Recent advances in humanoid robotics and embodied intelligence have enabled robots to perform increasingly complex manipulation tasks. However, musical instrument performance remains a formidable benchmark, demanding not only collision-free trajectory execution but also precise contact timing, asymmetric bimanual coordination, and target acoustic outcomes on physical instruments. The guqin, a seven-string fretless zither, presents unique manipulation challenges due to its millimetric string spacing, transient right-hand plucking, and sustained left-hand harmonic contacts. In this work, we present a physical heterogeneous dual-arm robotic system for phrase-level autonomous guqin performance. We formulate guqin playing as a hybrid discrete--continuous execution problem and develop a hierarchical planning framework that coordinates working-finger assignment, configuration continuity, obstacle avoidance, and tight bimanual contact schedules across consecutive musical events. The system integrates vision-guided instrument localization, tactile-based harmonic contact monitoring, and auditory-feedback-informed plucking parameter calibration. Real-world experiments on a 25-event phrase demonstrate that the system reliably executes coordinated open-string and seventh-hui harmonic sequences on a physical guqin, achieving 93.6\% and 96.8\% event correctness across repeated trials.
\end{abstract}

\section{Introduction}
Robotic musical performance serves as a rigorous testbed for robotic perception, dexterous manipulation, and physical interaction control \cite{bretan2016survey}. Unlike conventional pick-and-place or trajectory-following tasks, acoustic performance requires a robot to establish precise physical contact with an instrument under strict temporal constraints, where the functional success of manipulation is directly governed by physical and acoustic outcomes \cite{bir}.

The guqin is an ancient seven-string fretless plucked zither characterized by asymmetric bimanual interaction \cite{penttinen2006guqin}. During performance, the right hand executes transient plucking actions across densely spaced strings, while the left hand provides light, timed contact at specific inlaid markers (\textit{hui}) to produce harmonics or sustained contact for stopped notes. Achieving phrase-level robotic performance on a physical guqin introduces several fundamental challenges:

\textbf{1) High-dimensional bimanual coordination:} The system must schedule discrete musical events into collision-free, kinematic-feasible trajectories across two heterogeneous manipulators, ensuring configuration continuity across adjacent strings and avoiding singular or disjoint inverse-kinematics (IK) branches.

\textbf{2) Millimetre-level spatial and contact precision:} The closely spaced strings ($<$\,20\,mm) demand reliable 3D localization and fine positional offsets to avoid collateral damping or missed plucks.

\textbf{3) Tight inter-arm temporal synchronization:} Harmonics require the left hand to establish stable string contact strictly before right-hand excitation and release within a narrow temporal window after the pluck, necessitating real-time contact-state monitoring.
\begin{figure}[tbp]
    \centering
    \includegraphics[width=1\columnwidth]{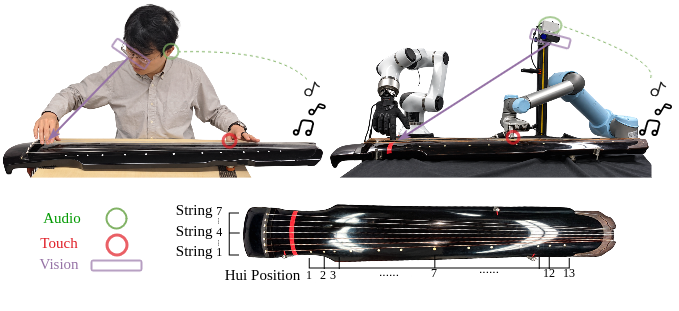}
 \caption{Physical heterogeneous dual-arm robotic system executing phrase-level performance on a real guqin. The setup integrates an Elfin manipulator with a multi-fingered hand for right-hand plucking, a UR5 manipulator with a contact-sensing hand for left-hand harmonic interaction, an RGB-D camera for instrument localization, and an acoustic feedback pipeline.}
    \label{fig:system_overview}
\vspace{-15pt}
\end{figure}

Prior research on stringed-instrument robotics has primarily focused on isolated plucking, bowing mechanisms, or single-contact explorations on instruments such as the guitar, cello, or guzheng \cite{yang2023guitarist, sudhoff2026cello,gorner2024pluck}. While recent studies have examined guqin acoustics, gestural typologies, and preliminary single-finger contact interfaces \cite{li2007guqin, zhang2026haptic}, realizing a complete, physical phrase-level guqin performance remains an open problem. Existing methods often rely on manual trajectory teaching or single-note triggering, lacking a unified framework to compile structured musical scores into feasible bimanual motions while monitoring physical and acoustic interaction states. 

To bridge this gap, this paper presents a complete heterogeneous dual-arm robotic system for phrase-level guqin playing on a real instrument (See Fig.~\ref{fig:system_overview}). Rather than focusing on specialized tactile transducer design or isolated single-note interactions, this work addresses the system-level problem of transforming structured multi-event phrases into coordinated, physically executable bimanual actions. The framework models the performance as an ordered set of hybrid discrete--continuous events, jointly optimizing fingering assignments, IK branch selections, trajectory smoothness, and bimanual temporal schedules. Furthermore, we incorporate multimodal perception into the execution pipeline: visual sensing provides 3D string and \textit{hui} geometry; tactile sensing gates left-hand harmonic contact states; and auditory feedback enables plucking parameter refinement during calibration intervals and event-level outcome evaluation during performance. 

To the best of our knowledge, this is the first physical robotic system capable of executing multi-event musical phrases with coordinated open-string and harmonic interactions on a real guqin.
The main contributions are as follows:
\begin{itemize}
    \item We formulate phrase-level guqin playing as a hybrid discrete--continuous task, mapping structured musical events to physically feasible dual-arm interaction targets.
    \item We develop a hierarchical bimanual planning and scheduling framework that resolves fingering assignment, IK branch continuity, obstacle avoidance, and tight plucking--harmonic contact timing.
    \item We deploy the system on a physical dual-arm robot with multimodal monitoring and calibration, validating repeated 25-event open-string and harmonic phrase execution on a real guqin.
\end{itemize}

\begin{figure*}[tbp]
\centering
\includegraphics[width=2\columnwidth]{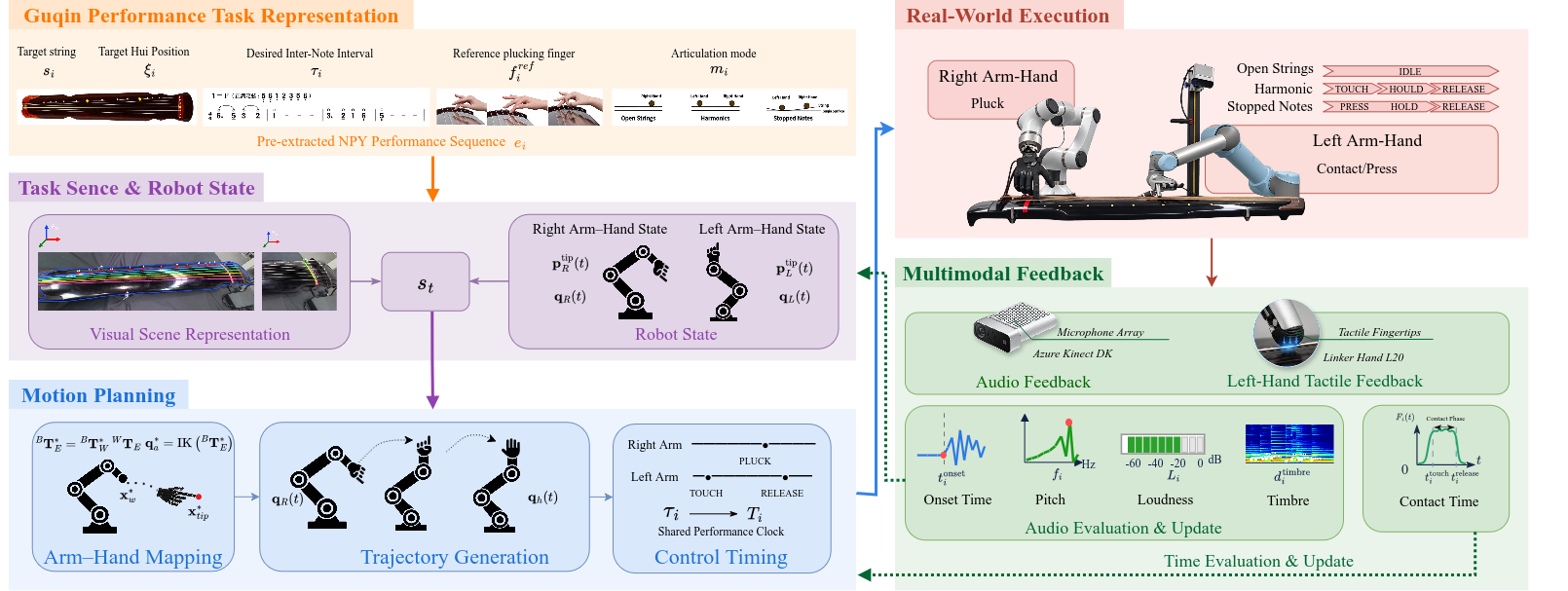}
\caption{Overview of the proposed phrase-level robotic guqin performance framework, integrating event-level score parsing, vision-guided 3D localization, hierarchical bimanual motion planning, and audio-tactile interaction monitoring.}
\label{fig:frame}
\vspace{-15pt}
\end{figure*}

\section{Related Work}
\subsection{Robotic String and Musical Instrument Performance}
Robotic musical performance has long served as a demanding benchmark combining high-dimensional actuation, precise kinematic planning, and multimodal physical interaction \cite{bretan2016survey}. Across various instruments, early systems primarily focused on reproducing kinematics or automated actuation mechanisms, such as robotic percussionists \cite{weinberg2006toward} and anthropomorphic flute-playing robots incorporating auditory parameter tuning \cite{solis2008waseda}. 

For keyboard and stringed instruments, research has increasingly emphasized dexterous contact and score-driven execution. Anatomically inspired and soft-skeleton hands have demonstrated piano playing with contact-sensitive dynamics \cite{zhang2011act, hughes2018piano}. For chordophones, Yang \textit{et al.} developed an expressive robotic guitarist capable of fret-shifting, strumming, and plucking on an acoustic guitar \cite{yang2023guitarist}. Recently, Sudhoff \textit{et al.} presented an end-to-end score-to-motion pipeline that converts MIDI scores into collision-aware bowing trajectories for robotic cello performance \cite{sudhoff2026cello}. These advancements illustrate a clear progression from single-string automation toward score-driven, collision-aware manipulation on physical instruments.

\subsection{Multimodal Perception and Learning in Instrument Playing}
Achieving reliable acoustic outcomes on physical instruments requires perceptual feedback to handle positioning uncertainties and contact dynamics. Learning-based frameworks such as RoboPianist \cite{zakka2023robopianist} and physical Sim2Real approaches \cite{zeulner2026piano} have explored high-dimensional dexterous control for keyboard playing. Beyond simulation, physical instrument manipulation heavily relies on tactile and auditory perception to evaluate contact states and acoustic responses. For instance, G{\"o}rner \textit{et al.} proposed a self-supervised exploration strategy combining audio-tactile interaction with Gaussian processes to localize strings and refine feasible plucking motions on a guzheng \cite{gorner2024pluck}. 

While these studies highlight the effectiveness of integrating tactile sensing and acoustic feedback to refine interaction primitives, most existing methods focus on single-arm setups or isolated contact skills. Extending multimodal feedback to asymmetric dual-arm systems executing long-horizon musical phrases requires coordinating discrete contact events with continuous multi-arm trajectories under strict temporal constraints.

\subsection{Guqin Performance and Robotic Interaction}
The guqin presents unique robotic manipulation challenges due to its fretless soundboard, close string arrangements, and asymmetric hand roles \cite{penttinen2006guqin}. Musicological and acoustic studies have analyzed characteristic guqin playing techniques, establishing typologies for sliding tones, pressed notes, and harmonics \cite{li2007guqin}. Computational approaches have further developed machine-learning models to recognize guqin fingering and modes from performance recordings \cite{huang2020guqin}.

In robotic guqin manipulation, prior research has largely focused on isolated string-contact phenomena or end-effector validation. For example, recent work evaluated a biomimetic tactile finger on isolated open-string plucks, harmonic parameter sweeps, and contact-triggered single-note coordination \cite{zhang2026haptic}. However, isolated single-finger or single-note interactions do not address the broader system-level challenges of phrase-level performance. Executing a continuous multi-event guqin phrase requires: (i) vision-guided 3D localization of dense strings and \textit{hui} positions; (ii) hierarchical bimanual planning to maintain configuration continuity across consecutive string switches; and (iii) tight temporal scheduling where left-hand harmonic contacts and right-hand plucks are monitored and calibrated via multimodal feedback. This paper presents an integrated system framework addressing these challenges on a physical dual-arm platform.

\section{System Framework And Methods}
\subsection{Task Representation and System Overview}
We formulate phrase-level physical guqin performance as a hybrid discrete--continuous manipulation task, where an ordered sequence of discrete musical events governs continuous bimanual coordination and localized string interactions. As illustrated in Fig.~\ref{fig:frame}, the overall framework integrates four core modules: (i)~\textit{Task Representation}, which translates structured score semantics into parametric performance events; (ii)~\textit{Multimodal Perception}, which establishes shared 3D geometric task frames and continuous state representations; (iii)~\textit{Hierarchical Bimanual Planning}, which resolves working-finger assignment, configuration continuity across strings, and inter-arm temporal synchronization; and (iv)~\textit{Execution and Monitoring}, which executes coordinated trajectories while leveraging tactile and auditory observations for contact gating and parameter refinement.

A reference musical phrase is formalized as an ordered sequence of $N$ discrete playing events:
\begin{equation}
    \mathcal{E} = \{e_i\}_{i=1}^N, \quad e_i = \left( s_i, \, \xi_i, \, \tau_i, \, f_i^{\mathrm{ref}}, \, m_i \right),
    \label{eq:event_representation}
\end{equation}
where $s_i \in \{1, \dots, 7\}$ denotes the target string index (numbered from player side outward), $\xi_i \in \{1, \dots, 13\}$ indicates the target \textit{hui} marker position along the soundboard, $\tau_i \in \mathbb{R}^+$ is the desired inter-onset interval (IOI) between consecutive notes, $f_i^{\mathrm{ref}}$ denotes the reference plucking finger from score notation, and $m_i \in \{0, 1, 2\}$ specifies the articulation mode corresponding to open-string tones, harmonics, and stopped notes, respectively.

Driven by this event stream, the framework bridges score semantics with continuous robotic execution through role-specific arm--hand coordination:
\begin{itemize}
    \item \textbf{Right-hand plucking system:} Excites target strings across the 7-string array near the first \textit{hui}. The planner assigns working fingers and selects smooth inverse-kinematics (IK) branches to ensure spatial reachability and posture continuity during rapid string switches.
    \item \textbf{Left-hand contact system:} Governed by articulation mode $m_i$. For open strings ($m_i=0$), the hand remains safely idle; for harmonics ($m_i=1$), it executes a timed \textit{Approach--Touch--Hold--Release} sequence at the target \textit{hui} $\xi_i$, where tactile sensing verifies contact onset before the right-hand pluck; for stopped notes ($m_i=2$), it firmly clamps the string against the soundboard throughout the sounding duration.
\end{itemize}

During execution, bimanual motions are synchronized via a unified performance clock anchored to plucking onsets. Proprioceptive feedback continuously gates phase transitions. Between performances or during dedicated calibration intervals, event-aligned tactile traces and auditory outcome metrics are evaluated against target acoustic templates to update local primitive offsets for subsequent phrase executions, establishing a structured multimodal feedback pipeline.

\begin{figure*}[tbp]
\centering
\includegraphics[width=2\columnwidth]{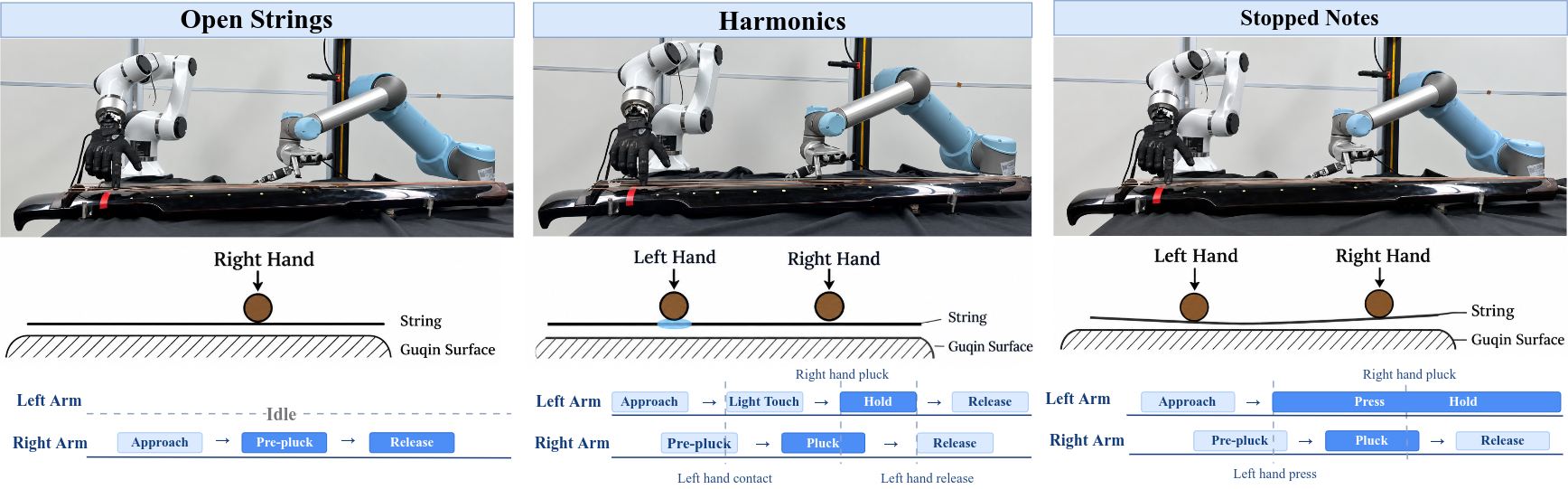}
\caption{State machine transitions and bimanual temporal coordination timeline. The top panels depict the phase state machines for right-hand plucking and left-hand harmonic interaction. The bottom timeline illustrates the unified performance schedule, detailing the left-hand contact window relative to the right-hand plucking onset $t^K_i$, the tactile contact-gating interval, and the post-pluck acoustic evaluation window.}
\label{fig:shixutu}
\vspace{-10pt}
\end{figure*}
\subsection{Vision-Based Spatial Representation and Target Generation}Precise guqin playing requires resolving sub-centimeter string intervals ($<$\,20\,mm) and converting musical semantics into metric manipulator targets. The vision pipeline first detects the guqin soundboard region via instance segmentation, and subsequently extracts the continuous centerlines of the seven strings and the discrete 3D coordinates of the thirteen \textit{hui} markers using a multi-task network with geometric line-fitting constraints.

Using the guqin coordinate frame $\mathcal{F}_G$ as the unified world reference for both arms, the visual observation is formalized as:
\begin{equation}
    z^V_t = \left( {}^C\mathbf{T}_G(t), \, \left\{ {}^G\boldsymbol{\gamma}_s(u) \right\}_{s=1}^7, \, \left\{ {}^G\mathbf{h}_\ell \right\}_{\ell=1}^{13} \right),
    \label{eq:visual_state}
\end{equation}
where ${}^C\mathbf{T}_G(t) \in \mathrm{SE}(3)$ denotes the guqin pose relative to the overhead camera, ${}^G\boldsymbol{\gamma}_s(u) \in \mathbb{R}^3$ describes the 3D centerline of string $s$ parameterized by the normalized longitudinal coordinate $u \in [0, 1]$, and ${}^G\mathbf{h}_\ell \in \mathbb{R}^3$ denotes the position of the $\ell$-th \textit{hui} marker.

For each string $s$, we establish a local orthonormal coordinate basis $\mathcal{R}_s = [\mathbf{t}_s, \, \mathbf{r}_s, \, \mathbf{n}_s]$:
\begin{equation}
    \mathbf{t}_s = \frac{\partial \boldsymbol{\gamma}_s / \partial u}{\|\partial \boldsymbol{\gamma}_s / \partial u\|}, \quad \mathbf{n}_s = \mathbf{n}_{\mathrm{board}}, \quad \mathbf{r}_s = \mathbf{n}_s \times \mathbf{t}_s,
\end{equation}
where $\mathbf{t}_s$, $\mathbf{r}_s$, and $\mathbf{n}_s$ represent the longitudinal string direction, the transverse in-plane direction toward the player, and the outward soundboard normal, respectively.

Given the performance event $e_i = (s_i, \xi_i, \tau_i, f_i^{\mathrm{ref}}, m_i)$, the task-space fingertip position targets in $\mathcal{F}_G$ are generated as:
\begin{equation}
    {}^G\mathbf{p}^a_i = {}^G\boldsymbol{\gamma}_{s_i}(u^a_i) + \mathcal{R}_{s_i} \boldsymbol{\delta}^a_i, \quad a \in \{R, L\},
    \label{eq:fingertip_target}
\end{equation}
where the right-hand longitudinal coordinate $u^R_i$ is anchored near the first \textit{hui} plucking zone, while the left-hand coordinate $u^L_i$ is aligned with the target \textit{hui} marker ${}^G\mathbf{h}_{\xi_i}$. The local offset vector $\boldsymbol{\delta}^a_i = [\delta^a_x, \delta^a_y, \delta^a_z]^T$ parameterizes the approach, pre-contact, plucking, and contact displacements for the active motion phase. 

Combining the fingertip target pose with hand kinematic transformations yields the desired wrist pose ${}^{B_a}\mathbf{T}_{W_a}$ in each manipulator's base frame $\mathcal{F}_{B_a}$:
\begin{equation}
    {}^{B_a}\mathbf{T}_{W_a} = {}^{B_a}\mathbf{T}_G \cdot {}^G\mathbf{T}_{F_a} \cdot \left( {}^{W_a}\mathbf{T}_{F_a}(\mathbf{q}_{H_a}) \right)^{-1},
    \label{eq:wrist_target}
\end{equation}
where ${}^{B_a}\mathbf{T}_G$ is obtained via eye-to-hand calibration, and $\mathbf{q}_{H_a}$ denotes the joint configuration of the dexterous hand.

\subsection{Hierarchical Bimanual Planning and Temporal Coordination}
The planner translates the event sequence $\mathcal{E}$ into collision-free, kinematic-feasible bimanual trajectories through three hierarchical stages: working-finger selection, continuous configuration optimization, and inter-arm temporal scheduling.

\subsubsection{Working-Finger Assignment}
To prevent kinematic awkwardness and frequent finger collisions across the densely arranged strings, the right-hand working finger sequence $\mathbf{f}^* = \{f_1^*, \dots, f_N^*\}$ is determined by solving an optimal assignment problem across all events:
\begin{equation}
    \mathbf{f}^* = \arg\min_{\mathbf{f}} \sum_{i=1}^N C_{\mathrm{ref}}(f_i, f_i^{\mathrm{ref}}) + \sum_{i=1}^{N-1} C_{\mathrm{tr}}(f_i, s_i, f_{i+1}, s_{i+1}),
    \label{eq:finger_opt}
\end{equation}
subject to $f_i \in \mathcal{F}(s_i)$, where $\mathcal{F}(s_i)$ denotes the set of kinematically feasible fingers for string $s_i$. The score-compliance cost is defined as $C_{\mathrm{ref}}(f_i, f_i^{\mathrm{ref}}) = w_r \mathbb{I}(f_i \neq f_i^{\mathrm{ref}})$, penalizing deviations from the reference score fingering. The transition cost $C_{\mathrm{tr}}$ is explicitly modeled as:
\begin{equation}
    C_{\mathrm{tr}} = w_s \mathbb{I}(f_i \neq f_{i+1}) + w_d |s_{i+1} - s_i|,
    \label{eq:transition_cost}
\end{equation}
which jointly penalizes frequent finger switches and large spatial string jumps between consecutive notes, ensuring smooth hand transitions.

\subsubsection{Configuration Continuity and Trajectory Generation}
For the assigned finger $f_i^*$, the right arm executes a phase state machine: $\text{Hover} \rightarrow \text{Pre-pluck} \rightarrow \text{Pluck} \rightarrow \text{Retract}$. The left arm concurrently plans an $\text{Approach} \rightarrow \text{Touch} \rightarrow \text{Hold} \rightarrow \text{Release}$ sequence for harmonics ($m_i=1$), or a sustained clamping trajectory for stopped notes ($m_i=2$).

To prevent discontinuous branch jumping in the 6-DoF inverse kinematics (IK) solution space across adjacent strings, candidate IK solutions $\mathcal{Q}_{f,s} = \{\mathbf{q}^{(1)}, \dots, \mathbf{q}^{(M)}\}$ are prioritized around calibrated nominal poses $\bar{\mathbf{q}}_{R,f,s}$. The full-phrase joint trajectory $\mathbf{Q}^*$ is optimized offline to minimize joint acceleration and jerk:
\begin{equation}
    \mathbf{Q}^* = \arg\min_{\mathbf{Q}} \sum_{i=1}^N \|\mathbf{q}_i - \bar{\mathbf{q}}_{f_i^*, s_i}\|^2_{\mathbf{W}} + \lambda_s \int \|\ddot{\mathbf{q}}(t)\|^2 dt,
\end{equation}
subject to joint position, velocity, and torque limits, as well as collision-free clearance between both arms and the guqin body.

\subsubsection{Bimanual Temporal Coordination}
As illustrated in Fig.~\ref{fig:shixutu}, bimanual execution is coordinated by a unified performance clock anchored to discrete plucking onsets $\{t^K_i\}_{i=1}^N$. The framework aligns the right-hand plucking phases ($\text{Hover} \rightarrow \text{Pre-pluck} \rightarrow \text{Pluck} \rightarrow \text{Retract}$) with the left-hand contact state machine ($\text{Approach} \rightarrow \text{Touch} \rightarrow \text{Hold} \rightarrow \text{Release}$), while enforcing strict temporal constraints for reliable sound production.
The inter-event scheduling enforces kinematic feasibility across consecutive events:
\begin{equation}
    t^K_{i+1} - t^K_i \ge \max \left( \kappa \tau_i, \, T^{\min}_{R,i}, \, T^{\min}_{L,i}, \, T^{\min}_{H,i} \right),
    \label{eq:temporal_feasibility}
\end{equation}
where $\kappa$ is a temporal scaling factor, and $T^{\min}_{a,i}$ is the minimum execution duration required by subsystem $a \in \{R, L, H\}$ to complete interaction at event $i$ and transition to event $i+1$.

For harmonic events ($m_i=1$), strict asymmetric contact constraints are enforced: the left hand must establish stable string contact strictly before right-hand plucking, and must release within a prescribed acoustic window after the pluck:
\begin{equation}
    t^{\mathrm{T,on}}_i < t^K_i, \quad \delta^{\min}_T \le t^{\mathrm{T,off}}_i - t^K_i \le \delta^{\max}_T,
    \label{eq:harmonic_timing}
\end{equation}
where $t^{\mathrm{T,on}}_i$ and $t^{\mathrm{T,off}}_i$ denote the contact onset and release instants, respectively. For stopped notes ($m_i=2$), the left hand completes string pressing before plucking and maintains contact throughout the sounding duration:
\begin{equation}
    t^{\mathrm{P,on}}_i < t^K_i, \quad t^{\mathrm{P,off}}_i - t^K_i \ge D_i,
    \label{eq:stopped_timing}
\end{equation}
where $D_i$ denotes the required post-pluck pressing duration. These constraints guarantee that multi-arm physical interactions satisfy the precise physical sound-production physics of each articulation mode.

\subsection{Multimodal Monitoring and Feedback-Informed Calibration}
To ensure robust physical interaction and acoustic sound production on a real guqin, the framework incorporates a two-stage multimodal pipeline: (i)~\textit{feedback-informed parameter calibration} prior to formal execution, and (ii)~\textit{event-level execution monitoring and contact gating} during phrase playing.

\subsubsection{Auditory-Feedback-Informed Primitive Calibration}
\label{sec:calibration}
Although vision provides structured 3D string geometry, millimetric discrepancies in string tension, fingertip compliance, and mounting errors can prevent clean acoustic excitation. Prior to formal performance, the system refines the nominal plucking targets for each working finger $f$ and string $s$ through an automated acoustic calibration procedure.

Starting from the visually estimated hover position $\mathbf{p}^H_{f,s}$, the local plucking offset $\boldsymbol{\eta}_{f,s} = [\Delta y_{f,s}, \, \Delta z_{f,s}]^T$ is iteratively adjusted in discrete steps ($\pm 1$\,mm). After each trial pluck, the Azure Kinect microphone records the acoustic response and extracts event features:
\begin{equation}
    \mathbf{z}^A = \left( t^A_{\mathrm{on}}, \, \hat{s}, \, \hat{f}_0, \, L, \, \mathrm{SNR} \right),
    \label{eq:acoustic_feature}
\end{equation}
comprising acoustic onset time $t^A_{\mathrm{on}}$, detected vibrating string $\hat{s}$, estimated fundamental frequency $\hat{f}_0$, loudness level $L$ (in dBFS), and signal-to-noise ratio $\mathrm{SNR}$. The acoustic discrepancy relative to a human reference sound template $\mathbf{z}^{A,\mathrm{ref}}_{f,s}$ is quantified by an acoustic loss:
\begin{equation}
    \mathcal{L}_A = w_s \mathbb{I}(\hat{s} \neq s) + w_f |\hat{f}_0 - f_0^{\mathrm{ref}}| + w_L |L - L^{\mathrm{ref}}|.
    \label{eq:acoustic_loss}
\end{equation}
A parameter update is accepted and committed only when $\mathcal{L}_A$ decreases, terminating when a prescribed similarity threshold is reached or a trial budget is exhausted. The calibrated offsets are stored in a primitive library $\Pi_{f,s}$ and held fixed during subsequent formal phrase performances.

\subsubsection{Event-Level Execution Monitoring and Contact Gating}
\begin{figure}[tbhp]
    \centering
    \includegraphics[width=\columnwidth]{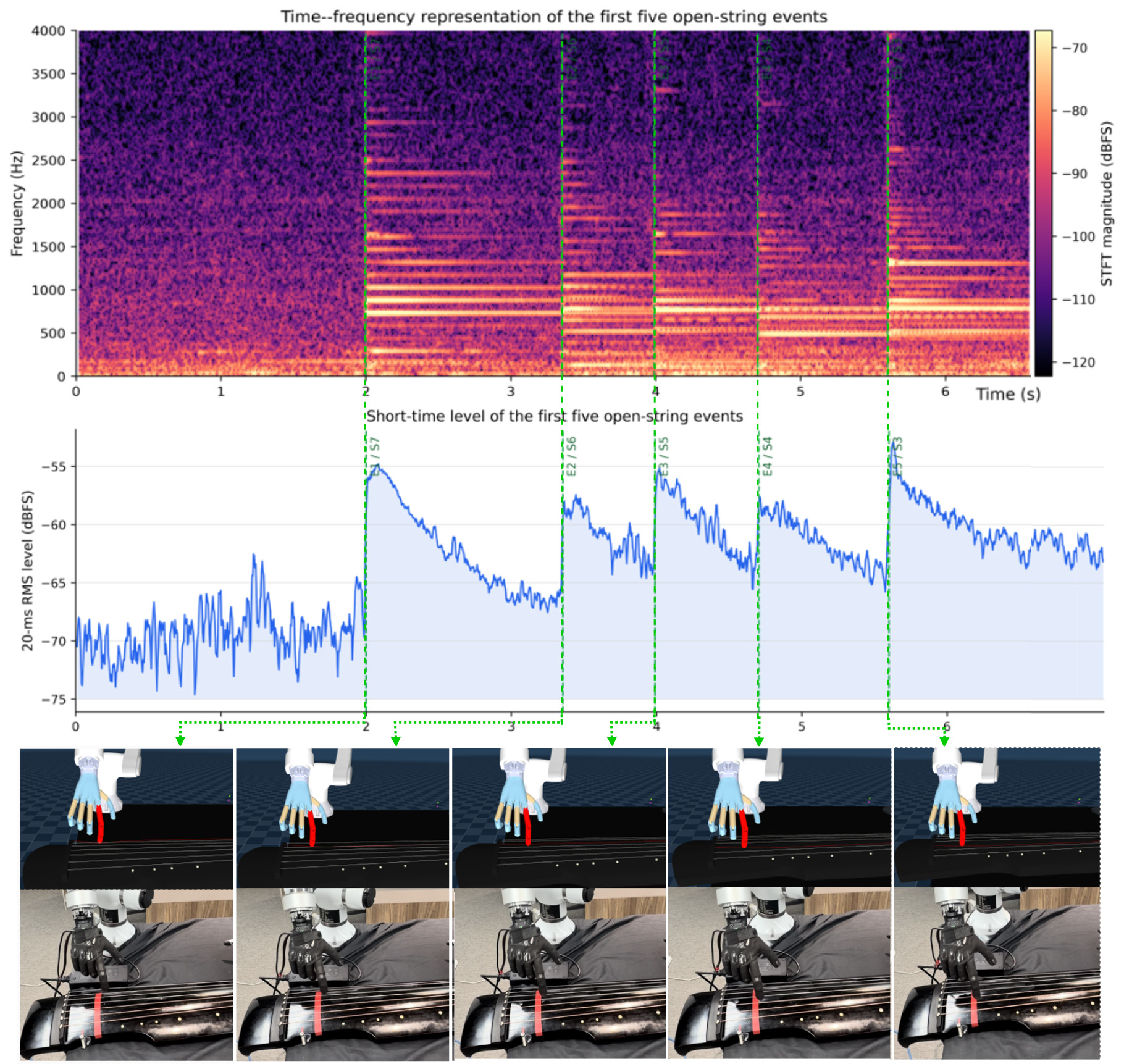}
\caption{Event-level acoustic monitoring of five consecutive open-string plucks. Top: STFT spectrogram and short-time RMS level time-aligned with commanded event onsets (dashed lines). Bottom: Synchronized snapshot sequences from simulation and physical robot execution.}
    \label{fig:acoustic_monitoring}
    \vspace{-10pt}
\end{figure}

During formal phrase execution, the system maintains strict trajectory continuity for rapid interaction phases (such as \textit{Pre-pluck}, \textit{Pluck}, and \textit{Touch}), while using multimodal observations for event-level gating and diagnostics:
\begin{itemize}
    \item \textbf{Proprioceptive tracking gating:} Joint-level tracking errors are continuously monitored. The controller enables the transition to the next active interaction phase only when manipulator and hand errors satisfy:
    \begin{equation}
        \|\mathbf{q}^a_{\mathrm{cmd}}(t) - \mathbf{q}^a_{\mathrm{meas}}(t)\|_{\infty} \le \boldsymbol{\epsilon}_a, \quad a \in \{R, L, H\}.
    \end{equation}
    \item \textbf{Tactile contact gating for harmonics:} For harmonic events ($m_i=1$), the left-hand tactile stream does not perform continuous trajectory servoing; instead, it serves as an event-gating signal within the temporal scheduler. The right-hand plucking motion is released only after left-hand contact establishment is validated:
    \begin{equation}
        \hat{t}^{\mathrm{T,on}}_i < t^K_i, \quad F_{\min} \le F_i(t^K_i) \le F_{\max},
    \end{equation}
    where $\hat{t}^{\mathrm{T,on}}_i$ is the measured contact instant, and $[F_{\min}, F_{\max}]$ defines the allowable normal contact range to avoid acoustic damping or string displacement.
    \item \textbf{Post-event outcome assessment:} Following each pluck, the auditory system verifies string correctness, fundamental frequency, and short-time acoustic energy. As illustrated in Fig.~\ref{fig:acoustic_monitoring}, continuous audio streams are time-aligned with commanded plucking onsets, showing clear STFT spectral harmonics and sharp RMS energy peaks corresponding to target string excitations. While committed trajectories within the active phrase are not altered mid-motion to preserve dynamic smoothness, anomalous events (e.g., missed plucks or misidentified strings) are logged to trigger targeted primitive recalibration during subsequent safe update intervals.
\end{itemize}

This structure establishes a robust multimodal pipeline linking visual spatial mapping, robot execution, tactile contact gating, and acoustic outcome evaluation.

\setcounter{dbltopnumber}{2}
\renewcommand{\dbltopfraction}{0.92}
\renewcommand{\dblfloatpagefraction}{0.75}
\renewcommand{\textfraction}{0.06}

\section{Experimental Results}\label{sec:experiments}
We evaluate the physical robotic system on three key aspects: 
\textbf{Q1:} How reliably does the system execute a 25-event guqin phrase across repeated open-string and seventh-hui harmonic (H7) trials?
\textbf{Q2:} How do multimodal observations guide offline parameter refinement and verify contact states? 
\textbf{Q3:} Can the event representation and planning framework transfer across different dexterous hands?

\begin{figure*}[tbhp]
    \centering
\includegraphics[width=0.8\textwidth]{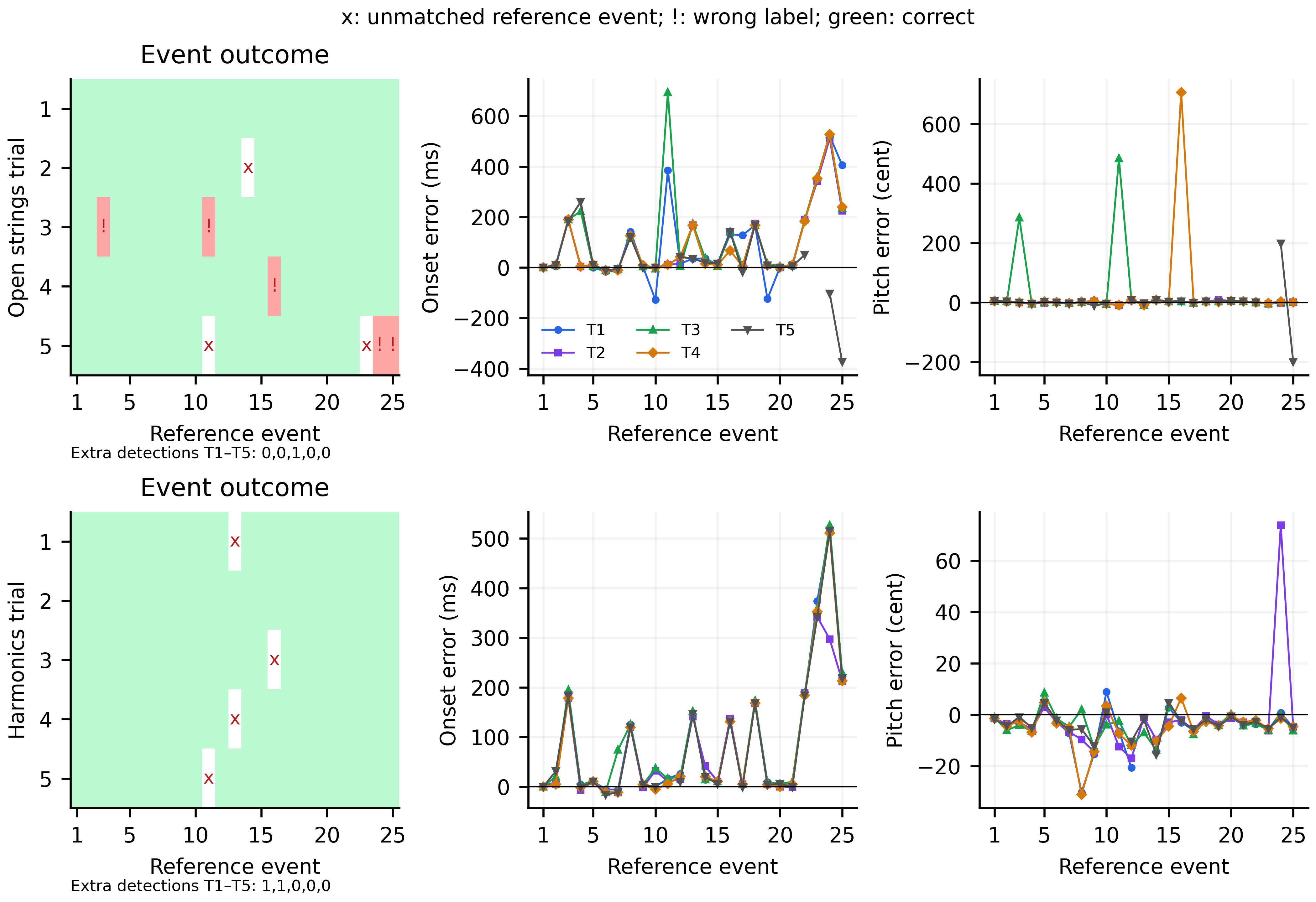}
\caption{Event-level outcomes, normalized onset errors, and pitch errors across five trials in open-string (top) and seventh-hui harmonic (H7, bottom) modes. Crosses indicate plucks whose sound level did not reach the detection threshold(M), exclamation marks indicate improper plucking motions that produced sounds inconsistent with the reference tones(W), and extra detections indicate that two strings were plucked simultaneously (E).}
\label{fig:abc_phrase}
\vspace{-10pt}
\end{figure*}

\subsection{Experimental Setup and Evaluation Protocol}
The experimental platform (Fig.~\ref{fig:system_overview}) consists of an Elfin E03 manipulator with a multi-fingered Wuji Hand for right-hand plucking, a UR5 manipulator with a Linker H L20 hand for left-hand harmonic interaction, an overhead Azure Kinect RGB-D camera for 3D localization, and a microphone recording at 48\,kHz \cite{bretan2016survey}. The system runs on Ubuntu 22.04 with ROS~2 Humble.

The benchmark task is a 25-event musical phrase extracted from the traditional guqin piece \textit{Cang Hai Xiao}. The phrase spans all seven strings over a 22.0\,s duration, requiring middle-finger plucks for events 6--10 and index-finger plucks for all others. We evaluate two performance modes using identical score timing: (i)~\textit{Open-string mode} ($m_i=0$), and (ii)~\textit{\textbf{Seventh-hui (H7)} harmonic mode} ($m_i=1$). For each mode, five consecutive trials were conducted using fixed calibrated parameters, with no recalibration between trials.

\subsubsection{Calibration Protocol}
Right-hand plucking targets are initialized from visual string localization and iteratively refined via acoustic loss $\mathcal{L}_A$ (Sec.~\ref{sec:calibration}). Left-hand harmonic contact positions are established by lowering the finger until raw tactile readings reach a stable baseline ($10$--$15$ raw units). All spatial offsets and hold durations remain fixed during formal performance trials.

\subsubsection{Acoustic Metrics}
Accepted acoustic onsets, string identities, and fundamental frequencies $\hat{f}_0$ are extracted from microphone streams. Performance is benchmarked against human-played reference templates. Each trial's first detected onset is aligned to the reference start without time scaling. A monotonic one-to-one temporal matching within a 0.25\,s correspondence window is applied. Reference templates and evaluation recordings were collected separately, and all trials were evaluated using the same fixed detector configuration. For each round of the experiment, we report:
\begin{itemize}
    \item \textbf{Event Correctness:} Ratio of correctly matched events with target string labels over 25.
    \item \textbf{Full-Phrase Success:} Strict binary metric requiring all 25 correct matches with zero extra detections.
    \item \textbf{Timing Errors:} Root-mean-square onset error $\mathrm{RMSE}_t$ and mean absolute inter-onset error $\mathrm{MAE}_{\mathrm{IOI}}$ across matched event pairs $\mathcal{P}$.
    \item \textbf{Pitch MAE:} $1200 |\log_2 (\hat{f}_0 / f_0^{\mathrm{ref}})|$ in cents across all valid matches.
    \item \textbf{Event Score:} $S_{\mathrm{evt}} = 10 \cdot \frac{2C}{25 + A}$, scaling the event-level F1-score to a 10-point scale ($C$: correct matches, $A$: total accepted detections).
\end{itemize}

\begin{table}[tbhp]
\centering
\caption{Five-Trial Acoustic Evaluation on the 25-Event Phrase. $M/W/E$ denotes Unmatched, Wrong-Label, and Extra Detections. Bold indicates best trial per mode.}
\label{tab:five_trial_summary}
\resizebox{\columnwidth}{!}{%
\begin{tabular}{cccccccc}
\hline
\textbf{Mode} & \textbf{Trial} & \textbf{Correct} & \textbf{M/W/E} & \textbf{$\mathrm{RMSE}_t$ (ms)} & \textbf{$\mathrm{MAE}_{\mathrm{IOI}}$ (ms)} & \textbf{Pitch MAE (cent)} & \textbf{$S_{\mathrm{evt}}$ / 10} \\ \hline
Open & O1 & \textbf{25/25} & \textbf{0/0/0} & 188.96 & 123.61 & 3.95 & \textbf{10.000} \\
Open & O2 & 24/25 & 1/0/0 & 157.26 & 101.33 & 3.64  & 9.796 \\
Open & O3 & 23/25 & 0/2/1 & 215.17 & 160.39 & 34.41 & 9.020 \\
Open & O4 & 24/25 & 0/1/0 & 156.04 & 93.94  & 31.84 & 9.600 \\
Open & O5 & 21/25 & 2/2/0 & 118.16 & 88.43  & 20.87 & 8.750 \\ \hline
\textbf{Open Avg.} & -- & \textbf{93.6\%} & 0.6/1/0.2 & $167.1 \pm 36.8$ & $113.5 \pm 29.4$ & $18.9 \pm 14.7$ & $9.43 \pm 0.52$ \\ \hline
H7   & H1 & 24/25 & 1/0/1 & 156.52 & 94.45  & 6.74  & 9.600 \\
H7   & H2 & \textbf{25/25} & 0/0/1 & \textbf{127.06} & \textbf{84.72} & 7.89  & \textbf{9.804} \\
H7   & H3 & 24/25 & 1/0/0 & 159.37 & 98.09  & 5.09  & 9.796 \\
H7   & H4 & 24/25 & 1/0/0 & 152.45 & 94.15  & 6.15  & 9.796 \\
H7   & H5 & 24/25 & 1/0/0 & 155.75 & 107.00 & \textbf{4.52} & 9.796 \\ \hline
\textbf{H7 Avg.}   & -- & \textbf{96.8\%} & 0.8/0/0.4 & $150.2 \pm 13.2$ & $95.7 \pm 8.0$   & $6.1 \pm 1.3$   & $9.76 \pm 0.09$ \\ \hline
\end{tabular}%
}
\vspace{-10pt}
\end{table}

\begin{figure*}[tbhp]
    \centering
    \includegraphics[width=\textwidth]{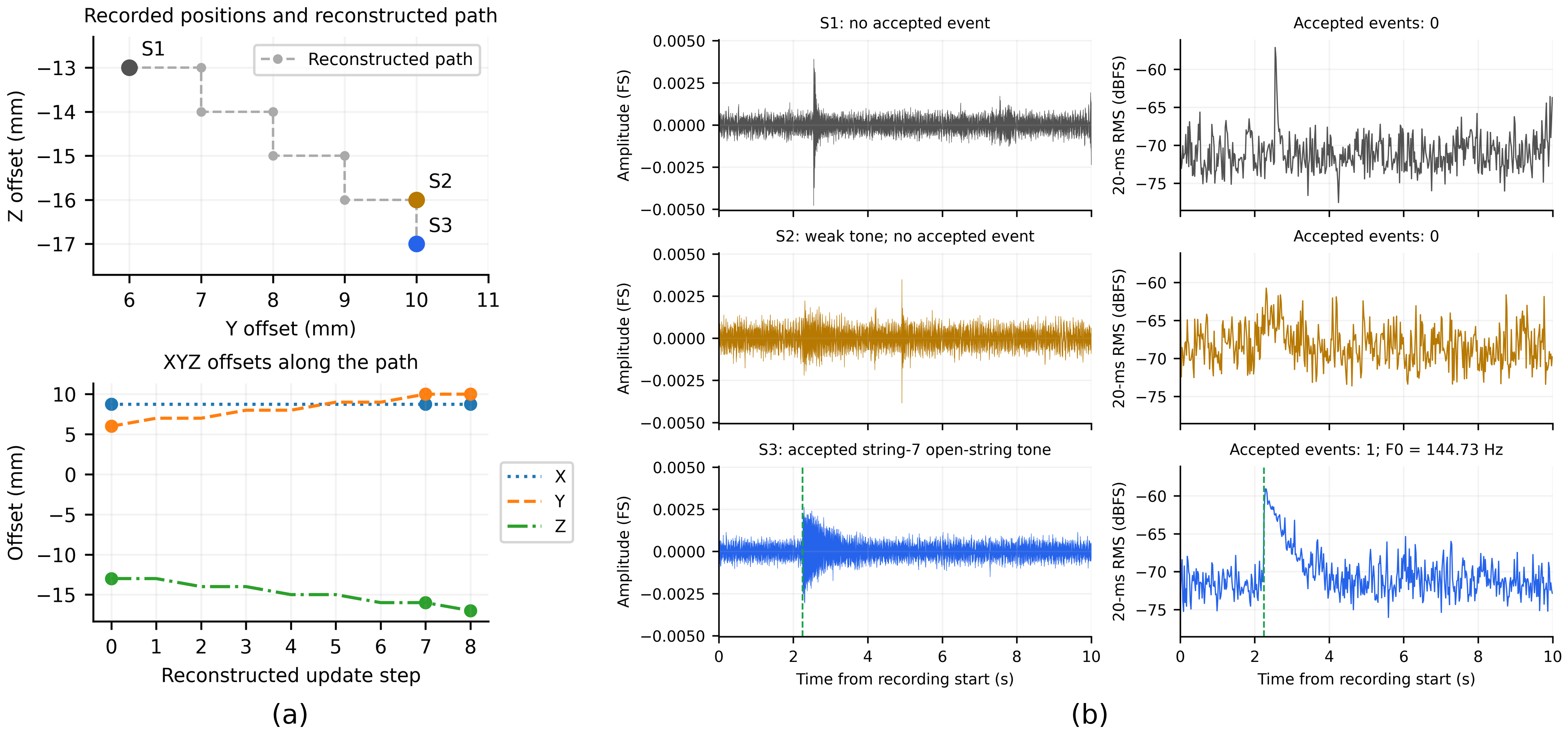}
    \caption{(a)~Right-hand positional calibration offsets for String 7. (b)~Waveforms and 20-ms RMS energy across calibration states, showing successful excitation at state S3.}
    \label{fig:calibration_acoustic}
\end{figure*}

\subsection{Phrase-Level Performance Validation (Q1)}
Table~\ref{tab:five_trial_summary} and Fig.~\ref{fig:abc_phrase} present the quantitative results across five repeated trials per mode. The system attains an aggregate event correctness of 93.6\% (117/125) for open strings and 96.8\% (121/125) for H7 harmonics. 

Under the strict full-phrase success metric, trial O1 achieved 25/25 correct notes with zero extra detections (1/5 success). For H7 harmonics, trial H2 successfully triggered all 25 target notes but logged one spurious extra detection (0/5 strict success). Notably, harmonic performance exhibits superior timing stability ($\mathrm{RMSE}_t = 150.2 \pm 13.2$\,ms) and pitch fidelity ($\mathrm{Pitch\ MAE} = 6.08 \pm 1.34$\,cents) compared to open strings ($18.94 \pm 14.73$\,cents). This confirms that left-hand contact effectively damps adjacent string resonance, yielding clearer fundamental onsets. 

\begin{figure*}[tbhp]
    \centering
    \includegraphics[width=0.9\textwidth]{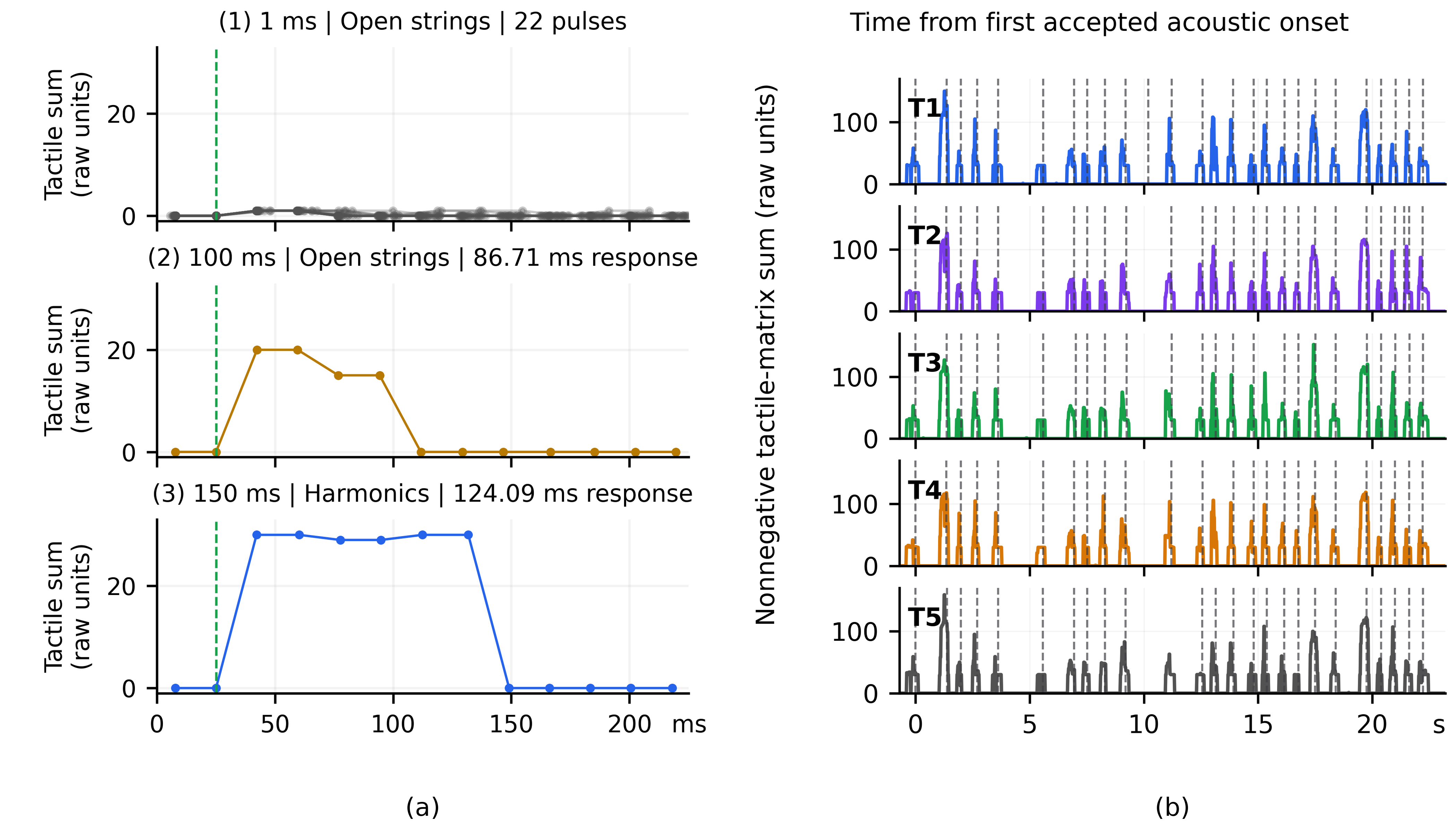}
    \caption{(a)~Tactile sensor responses under commanded hold times of 1, 100, and 150\, ms. (b)~Continuous tactile intensity traces across all five harmonics trials, with dashed lines marking verified acoustic onsets.}
    \label{fig:tactile_combined}
    \vspace{-10pt}
\end{figure*}

\subsection{Multimodal Observations during Calibration and Execution (Q2)}
We investigate how acoustic evaluation guides right-hand position refinement and how tactile sensing verifies left-hand harmonic gating.
\subsubsection{Right-Hand Acoustic Calibration}
Fig. \ref{fig:calibration_acoustic} (a) depicts the discrete update trajectory for string-7 plucking. While initial visual estimates (S1) failed to produce sufficient excitation, the acoustic-guided update reached state S3 ($\Delta y = 10$\,mm, $\Delta z = -17$\,mm), successfully triggering clean open-string vibration at $\hat{f}_0 = 144.7$\,Hz (Fig. \ref{fig:calibration_acoustic} (b)).

\subsubsection{Left-Hand Contact Gating and Hold Selection}
Fig. \ref{fig:tactile_combined} (a) compares tactile responses under different hold commands. A 1\,ms hold fails to establish sufficient acoustic boundary conditions (resulting in open-string tones), whereas hold durations of $100$--$170$\, ms (Table~\ref{tab:hold_commands}) provide the necessary contact damping to produce resonant harmonics. Fig. \ref{fig:tactile_combined} (b) confirms consistent tactile contact activity across all five H7 phrase executions, validating that the left hand established reliable string contact before each pluck.
\begin{table}[tbhp]
\centering
\caption{Calibrated Left-Hand Hold Durations (ms) by Plucking Finger and String. Dashes denote Unused Combinations.}
\label{tab:hold_commands}
\resizebox{0.85\columnwidth}{!}{%
\begin{tabular}{cccccccc}
\hline
\textbf{Finger / String} & \textbf{S1} & \textbf{S2} & \textbf{S3} & \textbf{S4} & \textbf{S5} & \textbf{S6} & \textbf{S7} \\ \hline
Right Index  & 150 & 150 & 170 & 125 & 100 & 150 & 150 \\
Right Middle & 100 & 100 & 150 & 100 & 100 & --  & --  \\ \hline
\end{tabular}%
}
\end{table}

\subsection{Cross-Hand Hardware Transfer Feasibility (Q3)}
\begin{figure}[tbhp]
    \centering
    \includegraphics[width=\columnwidth]{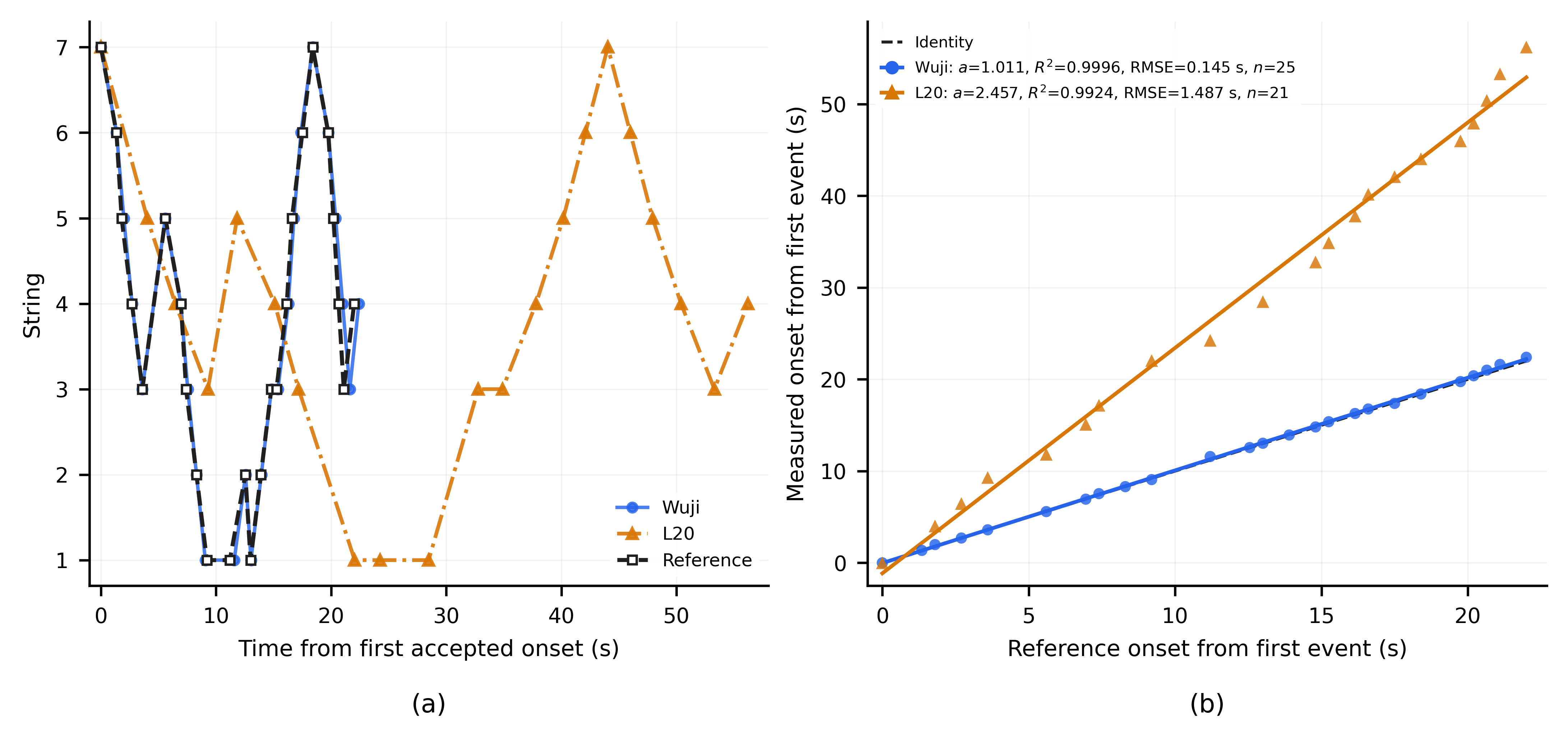}
    \caption{(a) Accepted event sequences. Discrete string event reproduction comparison between the Wuji Hand (22.4\,s) and replacement L20 Hand (56.2\,s) against the score reference. (b) Temporal transfer.}
    \label{fig:l20_transfer}
\end{figure}

To assess representation transferability, we replaced the right Wuji Hand with an L20 hand without altering high-level planning. Fig.~\ref{fig:l20_transfer} (a) compares the event sequences. The Wuji Hand executed the 25 events in 22.4\,s (100\% order coverage), while the L20 hand reproduced 21 events in 56.2\,s (84\% order coverage). All 21 accepted L20 events can be matched in reference order.
Fig.~\ref{fig:l20_transfer} (b) fits the order-aligned measured onsets as $\hat{t}=a t^{\mathrm{ref}}+b$. Wuji has $a=1.011$, $R^2=0.9996$, and a 0.145-s residual RMSE; L20 has $a=2.457$, $R^2=0.9924$, and a 1.487-s residual RMSE. Thus, the L20 execution preserves an approximately affine temporal progression while operating at a substantially expanded time scale.  While hardware differences in finger velocity limit dynamic equivalence, the result demonstrates that the task representation and inverse kinematics mappings can be transferred across different end-effectors.

\section{Conclusion}
This work presents a physical heterogeneous dual-arm robotic system for event-level guqin performance. The system converts a structured performance sequence into collision-aware, hardware-feasible bimanual motions through vision-based instrument localization, primitive selection, configuration continuity optimization, and temporal coordination. Tactile observations are used to monitor left-hand harmonic contact, while auditory observations support plucking-primitive calibration and event-level outcome assessment. Experiments on a 25-event phrase demonstrate repeated open-string and seventh-hui harmonic execution on a physical guqin.
\balance
\bibliographystyle{IEEEtran}
\bibliography{ref}
\end{CJK}
\end{document}